\documentclass{article}

\usepackage{PRIMEarxiv}

\usepackage[utf8]{inputenc} % allow utf-8 input
\usepackage[T1]{fontenc}    % use 8-bit T1 fonts
\usepackage{hyperref}       % hyperlinks
\usepackage{url}            % simple URL typesetting
\usepackage{booktabs}       % professional-quality tables
\usepackage{amsfonts}       % blackboard math symbols
\usepackage{nicefrac}       % compact symbols for 1/2, etc.
\usepackage{microtype}      % microtypography
\usepackage{lipsum}
\usepackage{fancyhdr}       % header
\usepackage{graphicx}       % graphics
\usepackage{multirow}
\usepackage{amsmath}
\usepackage{amssymb} 
\usepackage{tabularx}
\usepackage{makecell}
\usepackage{algorithm}
\usepackage{algpseudocode}
\usepackage{xcolor}
\usepackage{authblk}

\graphicspath{{media/}}     % organize your images and other figures under media/ folder

\title{Bayesian Mesh Optimization for Graph Neural Networks to Enhance Engineering Performance Prediction
}

\author[1,2]{Jangseop Park}
\author[1,2]{Namwoo Kang$^\dagger$}
\affil[1]{Cho Chun Shik Graduate School of Mobility, KAIST}
\affil[2]{Narnia Lab}

\begin{document}
\maketitle

\begingroup
    \renewcommand{\thefootnote}{$^\dagger$} % Change the footnote symbol for this entry
    \footnotetext{Corresponding author: nwkang@kaist.ac.kr}
\endgroup

\begin{abstract}
    In engineering design, surrogate models are widely employed to replace computationally expensive simulations by leveraging design variables and geometric parameters from computer-aided design (CAD) models. 
    However, these models often lose critical information when simplified to lower dimensions and face challenges in parameter definition, especially with the complex 3D shapes commonly found in industrial datasets. 
    To address these limitations, we propose a Bayesian graph neural network (GNN) framework for a 3D deep-learning-based surrogate model that predicts engineering performance by directly learning geometric features from CAD using mesh representation. 
    Our framework determines the optimal size of mesh elements through Bayesian optimization, resulting in a high-accuracy surrogate model. Additionally, it effectively handles the irregular and complex structures of 3D CADs, which differ significantly from the regular and uniform pixel structures of 2D images typically used in deep learning. 
    Experimental results demonstrate that the quality of the mesh significantly impacts the prediction accuracy of the surrogate model, with an optimally sized mesh achieving superior performance. 
    We compare the performance of models based on various 3D representations such as voxel, point cloud, and graph, and evaluate the computational costs of Monte Carlo simulation and Bayesian optimization methods to find the optimal mesh size. 
    We anticipate that our proposed framework has the potential to be applied to mesh-based simulations across various engineering fields, leveraging physics-based information commonly used in computer-aided engineering.
\end{abstract}

% keywords can be removed

%%%%%%%%%%%%%%%%%%%%%%%%%%% Introduction %%%%%%%%%%%%%%%%%%%%%%%%%%%%%%%

\section{Introduction}

Over the past few decades, computer simulations have become essential for modeling complex physics-based systems in various engineering applications, such as optimization design, uncertainty design, reliability analysis, and robust design. Despite their importance, these simulations are computationally intensive, as they aim to provide detailed representations of real-world systems. To address these computational challenges, surrogate models (SMs) and metamodels have become crucial in many engineering and scientific domains. These models offer efficient approximations of complex and computationally expensive simulations, significantly reducing computational costs while maintaining high accuracy \cite{cunningham2019investigation}. This efficiency is particularly valuable in iterative processes like design optimization. However, traditional surrogate modeling relies heavily on geometric parameters or design variables, which are critical for determining design shapes, sizes, and computational costs \cite{alizadeh2020managing}. This reliance presents challenges in representing complex 3D computer-aided design (CAD) datasets: low-dimensional representations often fail to capture the inherent geometric complexity of 3D shapes, while high-dimensional datasets are difficult to parameterize, affecting data resolution.

The growing complexity of engineering designs and the increasing reliance on 3D CAD models in modern engineering workflows further exacerbate these challenges. Surrogate models are essential in accelerating the design process by reducing the computational load of simulations, which is especially critical in industries like automotive, aerospace, and manufacturing. Traditional methods struggle with parameterizing high-dimensional 3D data and often face trade-offs between model accuracy and computational feasibility.

\begin{figure}[h]
    \centering\includegraphics[width=0.8\linewidth]{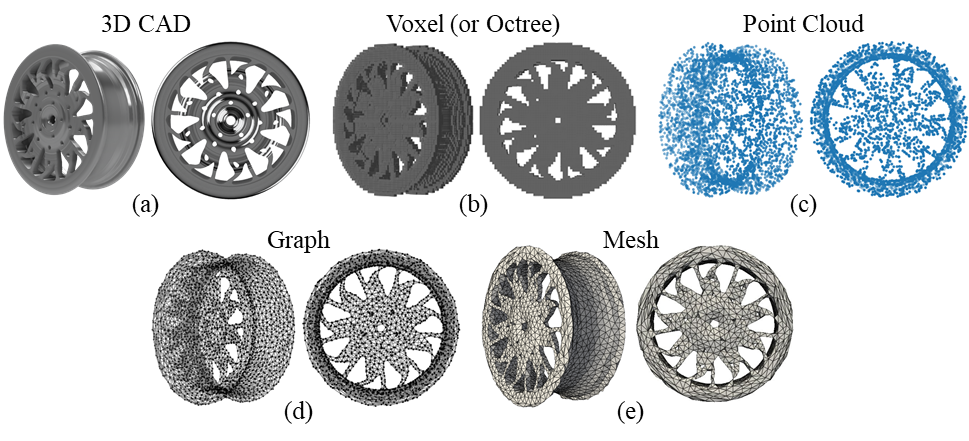}
    \caption{data representations. (a) 3D CAD. (b) Voxel (or Octree). (c) Point cloud. (d) Mesh. (e) Graph.}
    \label{fig:data-repre}
\end{figure}

Recent advancements in 2D computer vision with convolutional neural networks (CNNs) have driven significant progress in 3D tasks, including retrieval \cite{wu20153d, hanocka2019meshcnn, agathos20103d}, classification \cite{qi2017pointnet, qi2017pointnet++, wu20153d, xiang2015data, maturana2015voxnet, feng2019meshnet, hanocka2019meshcnn}, and segmentation \cite{tatarchenko2017octree, wang2017cnn, feng2019meshnet}. These advancements have significantly influenced the field of computer-aided design (CAD), where deep-learning-based surrogate models enhance performance prediction and design optimization by directly learning intrinsic 3D geometric structures. Various studies have employed different 3D data representations such as voxels, point clouds, meshes, and graphs to improve the accuracy of these surrogate models in representing 3D CAD as illustrated in Fig. \ref{fig:data-repre}.

For instance, studies on predicting engineering performance have used 3D CNN models with voxel representations \cite{williams2019design, yoo2021explainable, shin2023wheel}. Voxels, similar to 2D pixels but in 3D space, provide a structured approach but require significant memory to represent both occupied and non-occupied parts, complicating high-resolution modeling \cite{maturana2015voxnet, feng2019meshnet}. Point clouds, which are sets of unstructured 3D data points, are utilized in models like PointNet and PointNet++ \cite{qi2017pointnet, qi2017pointnet++}, offering a simpler but less information-dense representation compared to voxels and meshes. Meshes, combining vertices and faces, are popular for their efficiency in storing and rendering 3D data and their suitability for finite element analysis. Graph representations, which treat 3D models as nodes and edges, extend the benefits of meshes. Graph Neural Networks (GNNs) effectively learn from these representations, handling irregular data inputs and leveraging the connectivity information inherent in graphs. This makes GNNs particularly suitable for complex 3D CAD datasets \cite{scarselli2008graph}.

Furthermore, the accuracy and reliability of surrogate models heavily depend on the quality of input data. Preprocessing techniques are critical for enhancing model performance. Model-based data generation, as discussed by \cite{markovitch2021enhancing}, involves creating synthetic data to improve prediction accuracy and reliability. However, creating synthetic data that accurately captures the intricate details of 3D CAD models can be challenging, often leading to biases that compromise the surrogate model’s reliability. Active learning strategies, highlighted by \cite{balaprakash2013iterative} and \cite{granacher2021improvement}, optimize the training process by selectively querying the most informative data points, thereby reducing the amount of data needed. While effective, identifying informative samples in high-dimensional 3D data, which includes complex geometries, remains computationally expensive and challenging. Data augmentation, as described by \cite{jones2023data}, enhances model robustness by generating additional training data through transformations of existing data. Yet, in the context of 3D CAD models, simple transformations such as rotation or scaling might not preserve the essential geometric integrity, leading to suboptimal performance in surrogate models. Despite these advancements, determining the optimal mesh size for accurate surrogate modeling remains a significant challenge when dealing with 3D CAD datasets. Traditional methods often face a trade-off: an inappropriate mesh size can lead to either oversimplification, losing critical geometric details, or an excessive computational burden, which hampers model efficiency.

To address these challenges, we propose using graph representation for three primary reasons: it offers advantages in terms of memory efficiency and ease of rendering, is beneficial for implementing Graph Neural Networks (GNNs), which can handle the irregularity problem associated with varying input sizes of 3D CAD datasets, and aligns with methodologies that can potentially be applied to the Finite Element Method (FEM) in computer-aided engineering (CAE), suggesting its future applicability in solving partial differential equations (PDEs) robustly \cite{cao2020graph}.

In this study, we introduce a Bayesian GNN framework for a 3D deep learning-based surrogate model that predicts engineering performance with high accuracy, aiming to replace expensive simulations. The framework optimizes the trade-off between accuracy and mesh resolution efficiency by determining the optimal size of mesh elements (i.e., graphs) using Bayesian optimization (BO). Bayesian optimization is particularly advantageous as it systematically explores the mesh size parameter space and identifies the most suitable mesh resolution that balances accuracy and computational efficiency. This approach ensures that the surrogate model maintains high prediction accuracy without incurring unnecessary computational costs.

Our proposed framework can be applied to mesh-based simulations in various engineering fields. The contributions of this paper are as follows:

\begin{enumerate}
    \item \textbf{Graph Representation for 3D CAD Models:} We demonstrate the use of graph representation for 3D CAD models, highlighting its benefits in terms of memory efficiency, rendering capabilities, and managing irregular data inputs. This approach effectively addresses the limitations of traditional 3D representations such as voxels and point clouds.
    \item \textbf{High-Accuracy Surrogate Model for Engineering Performance Prediction:} Our framework provides a surrogate model that accurately predicts engineering performance, reducing the need for computationally expensive simulations. This model is particularly useful for iterative design processes and complex engineering analyses.
    \item \textbf{Efficiency of Bayesian Optimization:} We demonstrate the efficiency of Bayesian optimization in determining the optimal mesh element sizes. This method systematically explores the parameter space, significantly reducing computational costs compared to traditional methods like the Metropolis-Hastings algorithm of Markov Chain Monte Carlo (MCMC).
    \item \textbf{Applicability Across Engineering Disciplines:} Our proposed framework, which aligns graph representation with mesh representation, facilitates the future use of physical information derived from solving partial differential equations (PDEs) through the Finite Element Method (FEM). This capability enhances the framework’s ability to incorporate physics-based information, adapt to different simulation types, and thereby increase the utility of surrogate models in computer-aided engineering (CAE).
\end{enumerate}

By addressing the key challenges of memory efficiency, ease of rendering, and handling varying input sizes, our proposed method offers a significant advancement in the development of accurate and efficient surrogate models for complex engineering simulations. Moreover, the application of Bayesian optimization in determining the optimal mesh size ensures that the surrogate models are both accurate and computationally feasible, providing a robust solution to the limitations of existing methods.

The remainder of this paper is organized as follows. In Section \ref{sec:methodology}, we present the proposed framework, which consists of three stages: data preprocessing, model training, and Bayesian optimization. Section \ref{sec:results} presents the experimental results, including model performance, comparison with other models, and the relationship between mesh quality and prediction accuracy. Finally, Section \ref{sec:conclusion} concludes this paper.

%%%%%%%%%%%%%%%%%%%%%%%%%%% 2. Methodology %%%%%%%%%%%%%%%%%%%%%%%%%%%%%%%
%
\section{Methodology}
\label{sec:methodology}

\begin{figure*}[!t]
    \centering\includegraphics[width=1\linewidth]{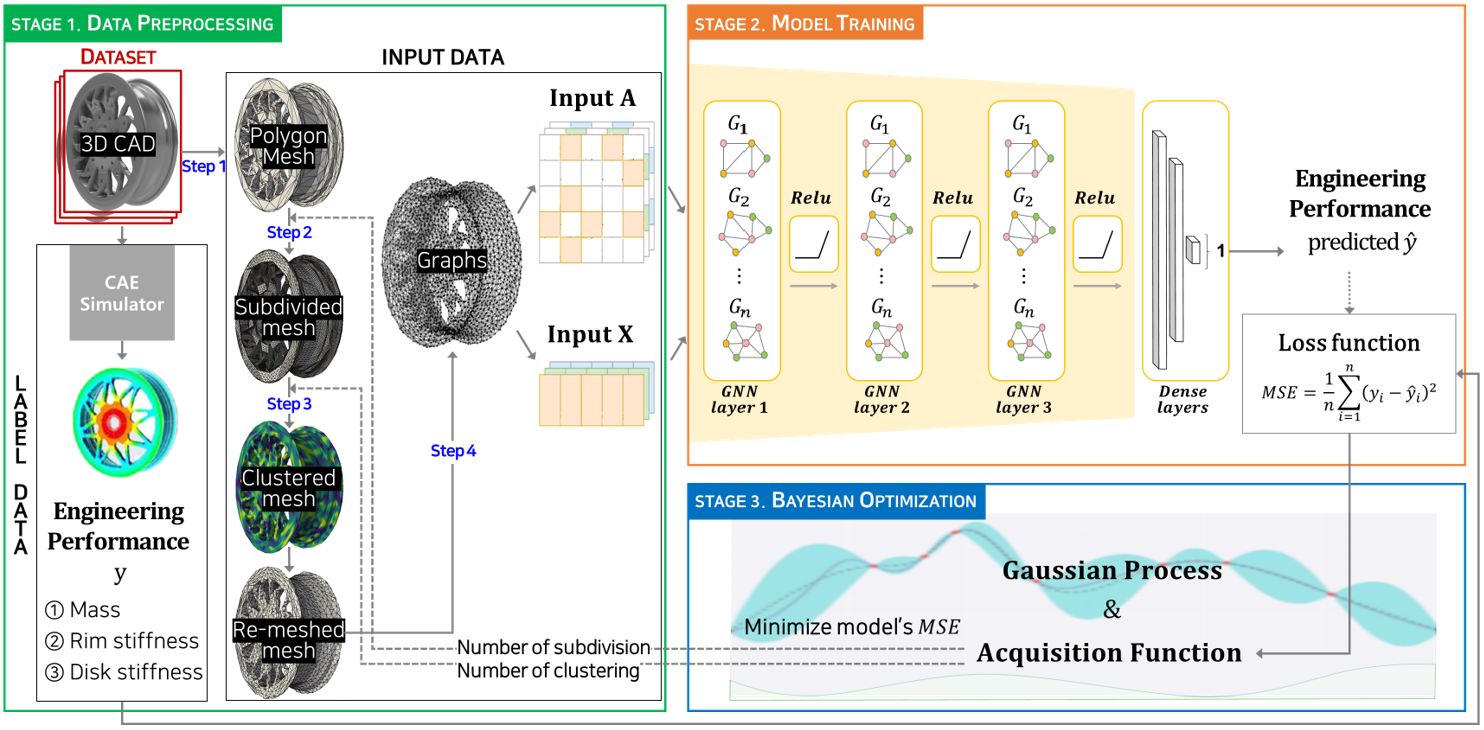}
    \caption{Proposed Bayesian GNN framework}
    \label{fig:GNN-framework}
\end{figure*}

We propose a Bayesian GNN framework utilizing a GNN model as a 3D DL-based surrogate model (SM) to predict engineering performance using a 3D CAD dataset, as illustrated in Fig. \ref{fig:GNN-framework}. This framework aims to derive the optimal mesh element size, thereby creating a high-accuracy surrogate model through Bayesian optimization.

\textbf{Stage 1.} includes data preprocessing of the label and input data. 3D CAD dataset were analyzed using modal analysis in the label data process to obtain the engineering performance. Four steps were performed to transform a 3D CAD dataset into a graph dataset during the input data process. In step 1, we convert the 3D CAD dataset into a polygon mesh dataset. In steps 2 to 3, we determine mesh element size with two hyperparameters related to subdividing and re-meshing the polygon mesh dataset. Finally, step 4 converts the re-meshed mesh into a graph as the input of the GNN model of stage 2.

\textbf{Stage 2.} is related to the GNN model for predicting the engineering performance. The GNN model was trained to minimize the mean square error (MSE) between the estimated and ground-truth values for the engineering performance. A detailed description of the GNN architecture, including the model hyperparameters, activation functions, dense layers, etc., is presented in Section \ref{subsec:stage2}.

\textbf{Stage 3.} is used to explore the dependent variable space (MSEs of the GNN model) while overlooking the independent variable space (two hyperparameters related to the size of the mesh elements). In other words, we use BO as a sequential strategy for the global optimization of the GNN model, known as a black-box and expensive-to-evaluate function. The acquisition function of BO recommends the two hyperparameters to be used in the next iteration, which conducts stages 1 to 3 sequentially. After several iterations until the predefined criteria are satisfied, we can finally obtain the optimal size of the mesh elements.
%
%%%%%%%%%%%%%%%%%%%%%%%%%%% 2.1. Data Preprocessing (stage 1) %%%%%%%%%%%%%%%%%%%%%%%%%%%%%%%
%
\subsection{Data Preprocessing (stage 1)}
\label{subsec:stage1}
%
%%%%%%%%%%%%%%%%%%%%%%%%%%% 2.1.1. Preprocessing of Label Data %%%%%%%%%%%%%%%%%%%%%%%%%%%%%%%
%
\subsubsection*{Preprocessing of Label Data}

This study utilized 925 3D road wheel CADs from the study of Yoo et
al. \cite{yoo2021integrating} as a 3D CAD dataset. The label data of the 3D CAD dataset consist of the engineering performance. Mode and frequency response analyses were performed on the 3D CAD dataset to obtain the rim and disk stiffness. For CAE automation, the macro function in Altairs Simlab \cite{AtairSimLab} was used to obtain $m,f,f_1$, and $f_2$.
\begin{equation}
    k_{rim}=\left(2\pi f\right)^2m
    \label{eq:k_rim}
\end{equation}
\begin{equation}
    k_{disk}={(2\pi f_2)}^2 \left[m-m\left(\frac{f_2^2}{f_1^2}\right)\right]
    \label{eq:k_disk}
\end{equation}
\begin{equation}
    y_{scale}^p=\frac{y^p-y_{min}^p\ \ }{y_{max}^p-y_{min}^p}
    \label{eq:y_scale}
\end{equation}

where $m$ corresponds to the value of the mass for each 3D wheel CAD, $f$ is the natural frequency in the normal mode of modal analysis, and $f_1$ and $f_2$ correspond to the resonance frequency and anti-resonance frequency in mode 11 (or the lateral mode) of the frequency response analysis, respectively. $p$ indicates the engineering performance: mass, disk stiffness, or rim stiffness. The rim stiffness and disk stiffness were calculated using Eq. \eqref{eq:k_rim} and Eq. \eqref{eq:k_disk}. Then, the mass, rim stiffness, and disk stiffness were used as the labels by min-max scaling (i.e., normalization), which adjusts the values to a fixed range (regularly 0 to 1) using Eq. \eqref{eq:y_scale}. 

\begin{figure*}[t]
    \centering\includegraphics[width=1\linewidth]{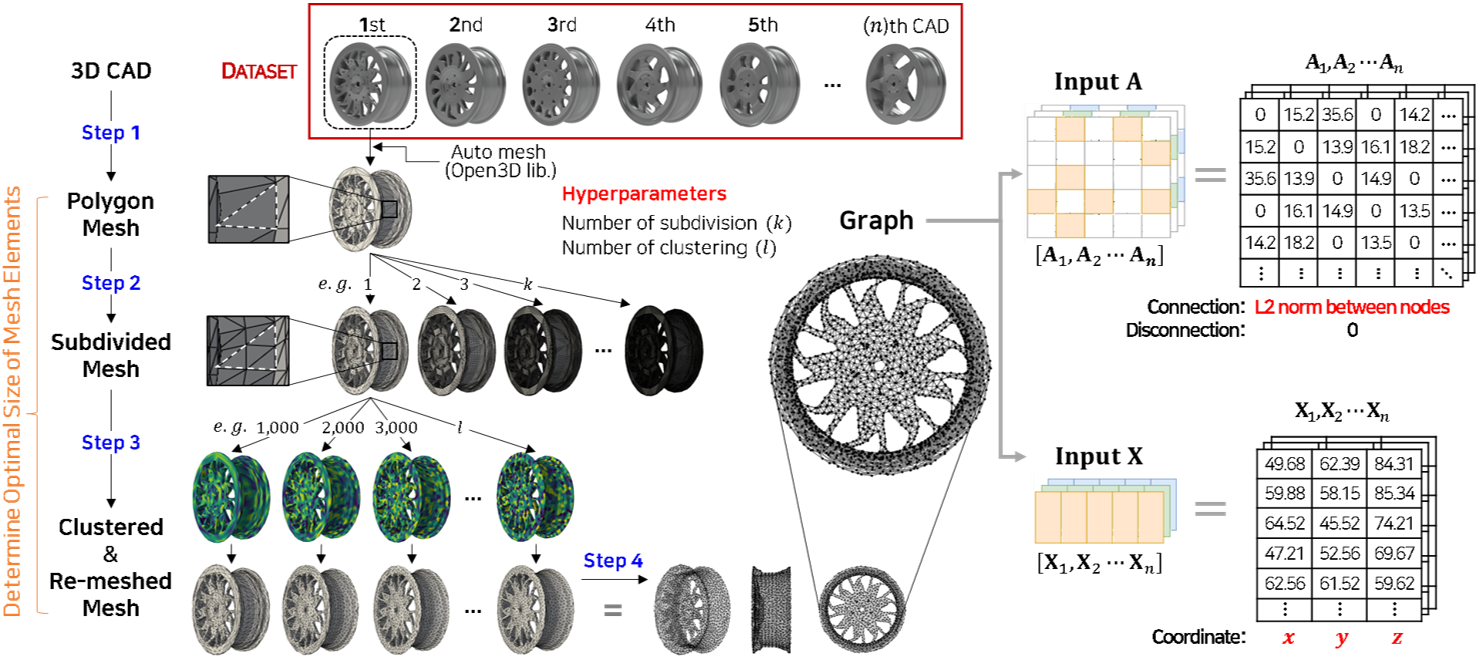}
    \caption{Preprocessing of input data comprises 4 steps}
    \label{fig:4steps}
\end{figure*}
\begin{figure}[h]
    \centering\includegraphics[width=0.55\linewidth]{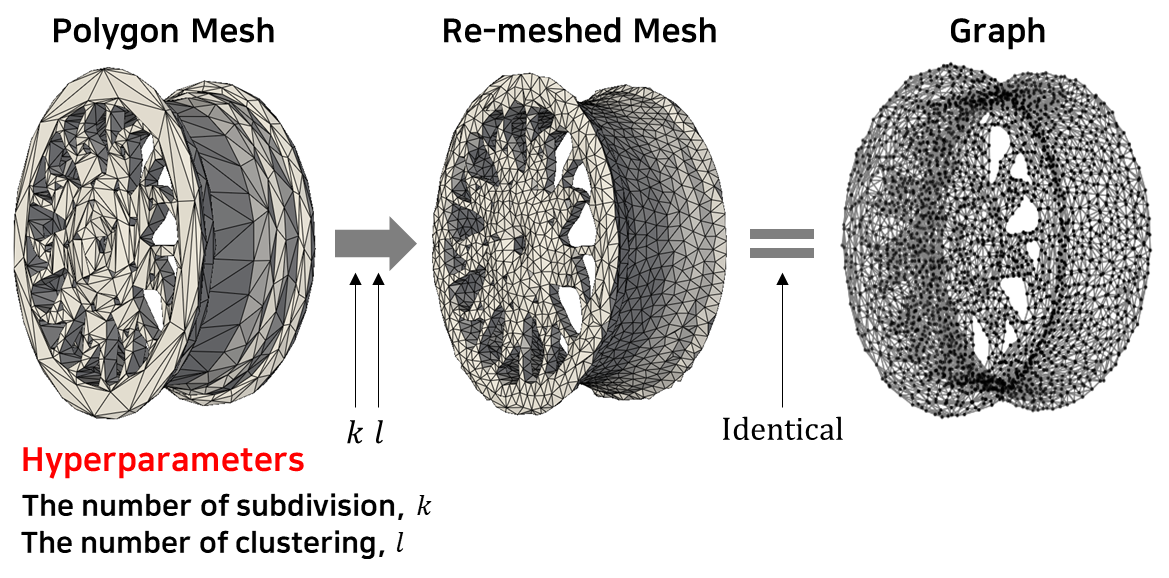}
    \caption{Summary for the re-meshing algorithm (steps 2 to 4) determined by two parameters k and l, and graph representation is identical to the mesh in terms of the data structure.}
    \label{fig:steps24}
\end{figure}
%
%
%%%%%%%%%%%%%%%%%%%%%%%%%%% 2.1.2. Preprocessing of Input Data %%%%%%%%%%%%%%%%%%%%%%%%%%%%%%%
\subsubsection*{Preprocessing of Input Data}
The preprocessing of input data for a 3D CAD dataset involves four steps: (1) polygon mesh conversion, (2) mesh subdivision, (3) clustering and re-meshing, and (4) graph transformation (see Fig. \ref{fig:4steps}). Steps 2 to 4 are summarized in Fig. \ref{fig:steps24}.

\begin{table}[h]
    \centering
    \caption{Polygon mesh dataset converted from CADs in step 1}
    \label{tab:Descrip of Graph}
    \begin{tabularx}{0.5\linewidth}{crrr}
        \toprule
         & Minimum & Maximum & Average \\
         \midrule
         Number of nodes & 624.0 & 1,692.0 & 1,039.0 \\
         Number of edges & 1,932.0 & 5,148.0 & 3,271.0 \\ 
         Number of faces & 1,287.0 & 4,238.0 & 2,177.3 \\ 
        \bottomrule
    \end{tabularx}
    \label{tab:my_label}
\end{table}
\textbf{Step 1:} The collected 3D CAD dataset is converted into a polygon mesh dataset with triangular meshes, which consists of a combination of nodes (or vertices) and faces. The nodes include a connectivity list that describes their connections. Open3D library with Python API \cite{Zhou2018} is used to automatically convert the 3D CAD dataset into a polygon mesh dataset with irregular mesh element size as show in Table \ref{tab:Descrip of Graph}.

\textbf{Step 2:} We utilize the re-meshing algorithm in steps 2–3 with anisotropic discrete Voronoi diagrams (DCVD) \cite{valette2008generic}. DCVD has two advantages: handling complex models with several million triangles and preserving the features of general objects \cite{khan2020surface}. Step 2 involves subdividing the meshes to increase the number of meshes when polygon meshes converted from 3D CAD data have fewer nodes. The number of subdivisions $k$ as the first hyperparameter implies that one triangular mesh is divided into $4^k$ triangular meshes. For example, if $k$ is 1, one triangular mesh is subdivided into four triangular meshes (white dashed lines in Fig. \ref{fig:4steps}).

\textbf{Step 3:} The DCVD algorithm minimizes the global energy term by partitioning (or clustering) the input meshes to efficiently distribute the node budget by the number of clustering $l$ as the second hyperparameter. DCVD is a theoretical clustering algorithm for preserving high-quality meshes based on the duality between the Delaunay triangulation (DT) and Voronoi diagrams (VD) which is widely used in geometrics. The clustered mesh can be easily transformed into a re-meshed mesh because the re-meshed mesh (DT) is a straight-line dual of the clustered mesh (VD) \cite{aurenhammer2013voronoi}. Steps 2 to 3 can be summarized with the re-meshing algorithm by determining two parameters, $k$ and $l$.

\textbf{Step 4:} The mesh representation is converted into a graph representation as the input form of the GNN model. Re-meshed meshes can be transformed into graphs whose nodes and edges are the same as those of the mesh \cite{fey2018splinecnn}. The details of the mesh-to-graph conversion are introduced in Section \ref{subsec:stage2}.

%
%%%%%%%%%%%%%%%%%%%%%%%%%%% 2.2. Model Training (stage 2) %%%%%%%%%%%%%%%%%%%%%%%%%%%%%%%
\subsection{Model Training (stage 2)}
\label{subsec:stage2}
%%%%%%%%%%%%%%%%%%%%%%%%%%% 2.2.1. Graph Dataset %%%%%%%%%%%%%%%%%%%%%%%%%%%%%%%
\subsubsection*{Graph Dataset}

We converted the 3D CAD dataset, which consists of 925 3D CAD models, into a re-meshed mesh dataset through steps 1 to 3 of the input data preprocessing. Let $\mathcal{M}$ be the re-meshed dataset as $\mathcal{M}\equiv\left\{M_1,M_2\ldots M_{925}\right\}$. Here, we represent a mesh data $M$ as $M=\{\mathcal{V},\mathcal{F}\}$ with the node set $\mathcal{V}$ and face (or cell) set $\mathcal{F}$. In step 4 of Section \ref{subsec:stage1}, one mesh data $M$ is transformed into the graph data $G$ as $G=\{\mathcal{V},\ \mathcal{E}\}$ with node set $\mathcal{V}$ and edge set $\mathcal{E}$, which is identical to the mesh in terms of the data structure. Finally, we obtain the graph dataset $\mathcal{G}\equiv\left\{G_1,G_2\ldots G_{925}\right\}$. 

One graph $G$ as the input data of a GNN model is represented in two matrix forms with adjacency matrix $\mathbf{A}\in\mathbb{R}^{N\times N}$ and node feature matrix $\mathbf{X}\in\mathbb{R}^{N\times F}$, where $N$ is the number of nodes and $F$ is the number of node features. The adjacency matrix $\mathbf{A}$ includes edge information about the relationships between adjacent nodes. Equally, a node feature matrix $\mathbf{X}$ includes the node coordinates $(x,\ y,\ z)$. 

One engineering performance label $y^p$ is obtained from one 3D CAD by performing CAE simulations. Given a set of datasets $\mathcal{D}^p=\left\{\left(G_i,y_i^p\right)\right\}\ i=1,\ldots,925$, for each engineering performance, the goal of a GNN model as a DL-based surrogate model is to learn the relationships $f:\ \mathcal{G}\rightarrow\mathcal{Y}$, where $y_i^p\in\mathcal{Y}^p$ is the label corresponding to graph $G_i\in\mathcal{G}$.

Although we adopt the GNN model \cite{bianchi2020spectral} as the baseline for our proposed framework, we compare the performance of both GNN and GCN models without re-meshing in Section \ref{subsec:comparison-different-models}.

\subsubsection*{Graph Neural Networks}

Graph Neural Networks (GNNs) are designed to work directly with graph-structured data. The key idea behind GNNs is the concept of message passing, where nodes in the graph exchange information with their neighbors to generate node embeddings that capture both local and global graph structures.

The adjacency matrix $\mathbf{A}\in\mathbb{R}^{N\times N}$ of the GNN model represents the connectivity between nodes, with entries that can be binary values indicating the presence or absence of an edge or weights representing the strength of connections. In this study, the adjacency matrix contains either connective or non-connective values: L2 norm or 0, as shown in Fig. \ref{fig:4steps}.

The propagation rule in a GNN layer can be described by the following general equation:

\begin{equation}
\mathbf{H}^{(l+1)} = \sigma \left( \mathbf{A} \mathbf{H}^{(l)} \mathbf{W}^{(l)} \right)
\label{eq
}
\end{equation}

where $\mathbf{H}^{(l)}$ is the node feature matrix at the $l$-th layer. $\mathbf{A}$ is the adjacency matrix representing the connection weights between nodes. $\mathbf{W}^{(l)}$ is the trainable weight matrix at the $l$-th layer, and $\sigma$ is a non-linear activation function, commonly ReLU (Rectified Linear Unit). 

The initial node features are given by $\mathbf{H}^{(0)} = \mathbf{X}$. The output of each layer is a set of node embeddings that have aggregated information from their neighbors through the adjacency matrix.

\subsubsection*{Graph Convolutional Networks}

Graph Convolutional Networks (GCNs) are a specific type of GNN that perform graph convolutions analogous to the convolutions used in Convolutional Neural Networks (CNNs) for image data. GCNs extend the idea of convolution to graph-structured data by redefining the convolution operation in the spectral domain. In this study, the adjacency matrix $\mathbf{A}$ of the GCN contains connective or non-connective values: 1 or 0.

The adjacency matrix $\hat{\mathbf{A}}\in\mathbb{R}^{N\times N}$ of the GCN includes self-loops, ensuring that a node’s features are also considered in its update. This is represented as $\hat{\mathbf{A}} = \mathbf{A} + \mathbf{I}$, where $\mathbf{I}\in\mathbb{R}^{N\times N}$ is the identity matrix. The GCN model normalizes this adjacency matrix using the degree matrix $\hat{\mathbf{D}}$, resulting in $\hat{\mathbf{D}}^{-\frac{1}{2}}\hat{\mathbf{A}}\hat{\mathbf{D}}^{-\frac{1}{2}}$.

The propagation rule for a GCN layer is given by:

\begin{equation}
\mathbf{H}^{(l+1)} = \sigma \left( \hat{\mathbf{D}}^{-\frac{1}{2}}\hat{\mathbf{A}}\hat{\mathbf{D}}^{-\frac{1}{2}}\mathbf{H}^{(l)}\mathbf{W}^{(l)} \right)
\label{eq
}
\end{equation}

where $\mathbf{W}^{(l)}\in\mathbb{R}^{F\times d}$ is the trainable weight matrix, and $\mathbf{H}^{(l+1)}$ represents the node embeddings at layer $l+1$. Similar to GNNs, $\mathbf{H}^{(0)} = \mathbf{X}$ is the initial node feature matrix.

\subsubsection*{Comparison of GNN and GCN}

The key difference between GNNs and GCNs is the structure of the adjacency matrix $\mathbf{A}$ representing the edge connectivity between nodes. GNNs typically use a weighted adjacency matrix that can include various types of edge weights, while GCNs use a normalized adjacency matrix with self-loops. Both GNNs and GCNs operate on graph-structured data but differ in their specific approaches to aggregating information from neighboring nodes. GNNs use a general message-passing framework, whereas GCNs apply a spectral convolution approach with normalized adjacency matrices. We compare the performance of both GNN and GCN models without re-meshing in Section \ref{subsec:comparison-different-models}.

Both models use the same node feature matrix $\mathbf{X}$, normalized by the following equation:

\begin{equation}
\mathbf{X}_{\text{scale}} = \frac{\mathbf{X} - \mathbf{X}_{\min}}{\mathbf{X}_{\max} - \mathbf{X}_{\min}}
\label{eq
}
\end{equation}

This normalization ensures that the input features are scaled appropriately for training.

%
%%%%%%%%%%%%%%%%%%%%%%%%%%% 2.2.4. Model Architecture %%%%%%%%%%%%%%%%%%%%%%%%%%%%%%%
%
\subsubsection*{Model Architecture} 

The proposed GNN model architecture aims to predict the engineering performance obtained from CAE. The graph layers serve as graph feature extractions, and dense (or fully connected) layers are used for regression. There are three common numbers for both the graph and dense layers. The graph layers have a fixed latent vector size of 512 dimensions. The dense layers have latent vector sizes of 500, 200, and 25 dimensions, sequentially. A rectified linear unit (ReLU) was used as the activation function, except for the last layer. The proposed GNN model was trained using the Adam optimizer with a learning rate of 0.0002 and a batch size of 1. The epochs were set to 10,000, but the early stopping of callback techniques was used to prevent overfitting. The patience of the early stopping which is used to determine the duration of model training was set to 50 for validation loss. The loss function was set as the mean squared error (MSE) to predict each engineering performance, calculated using Eq. \eqref{eq:MSE}. A total of 925 (100$\%$) graphs were split into 740 (80$\%$) training, 92 (10$\%$) validation, and 93 (10$\%$) test sets. The labels of the validation set were not used for training. 
\begin{equation}
    MSE^p=\frac{1}{n}\sum_{i=1}^{n}{{(y}_i^p-{\hat{y}}_i^p)}^2
    \label{eq:MSE}
\end{equation} where $n$ is the number of graphs $G_i\ (G_i\in\mathcal{G})$ and $y_i^p$ is the $\left(i\right)$th ground label for each engineering performance according to the $\left(i\right)$th graph $G_i$.

%%%%%%%%%%%%%%%%%%%%%%%%%%% 2.3. Bayesian Optimization (stage 3) %%%%%%%%%%%%%%%%%%%%%%%%%%%%%%%
%

\subsection{Bayesian Optimization (stage 3)}
\label{subsec:stage3}

\begin{algorithm}
\caption{Bayesian Optimization (BO) with Expected Improvement (EI) for GNN Model}
\label{alg:BO-GNN}
\begin{algorithmic}[1]
\State \textbf{Input:} 3D CAD dataset, $t_{\text{max}}$, $p_{\text{max}}$
\State \textbf{Output:} Optimal hyperparameters $(k^*, l^*)$
\State \textbf{Initialize:} 
\State \hspace{1cm} $t \gets 1$ \Comment{Current iteration for Bayesian optimization}
\State \hspace{1cm} $p \gets 1$ \Comment{Current patience count for Bayesian optimization}
\State \hspace{1cm} $MSE_{\text{best}} \gets \infty$ \Comment{Best MSE so far}
\State \hspace{1cm} Set bounds for $k$ and $l$

\State Convert 3D CAD dataset to polygon mesh \textcolor{blue}{(Step 1)}

\While{$t \leq t_{\text{max}}$ and $p \leq p_{\text{max}}$}
    \If{$t = 1$}
        \State Initialize the surrogate model of BO (e.g., Gaussian Process) with initial history $\mathcal{H}$
    \EndIf

    \State Select hyperparameters $(k_t, l_t)$ by maximizing EI:
    \[
    (k_t, l_t) = \arg \max_{(k, l)} \text{EI}(k, l; \mathcal{H}_{t-1})
    \]

    \State Perform mesh subdivision with $k_t$ \textcolor{blue}{(Step 2)}
    \State Apply clustering and re-meshing with $l_t$ \textcolor{blue}{(Step 3)}
    \State Convert re-meshed mesh to graph representation \textcolor{blue}{(Step 4)}

    \State Initialize GNN model weights
    \State Train the GNN model until early stopping
    \State $MSE_t \gets$ Evaluate MSE of the GNN model

    \If{$MSE_t < MSE_{\text{best}}$}
        \State $MSE_{\text{best}} \gets MSE_t$
        \State $p \gets 1$ \Comment{Reset patience count}
    \Else
        \State $p \gets p + 1$ \Comment{Increment patience count}
    \EndIf

    \State Augment history $\mathcal{H}_t = \mathcal{H}_{t-1} \cup \{((k_t, l_t), MSE_t)\}$
    \State Update the surrogate model with $\mathcal{H}_t$
    \State $t \gets t + 1$
\EndWhile

\State \Return $(k^*, l^*) = \arg \min_{(k, l) \in \mathcal{H}} \text{MSE}(k, l)$
\end{algorithmic}
\end{algorithm}

Bayesian Optimization (BO) has been widely used in machine learning \cite{snoek2012practical} to find hyperparameters that optimize the performance of DL models. BO is employed in our framework to determine the optimal hyperparameters for mesh element size, specifically the number of subdivisions \(k\) and the number of clusters \(l\). This process ensures that the Graph Neural Network (GNN) model achieves high prediction accuracy while maintaining computational efficiency. Below, we describe the detailed procedure of the algorithm used for BO with Expected Improvement (EI).

Firstly, the input to the algorithm includes the 3D CAD dataset, the maximum number of iterations for Bayesian optimization (\(t_{\text{max}}\)), and the maximum patience count (\(p_{\text{max}}\)). The goal is to output the optimal hyperparameters \((k^*, l^*)\) that minimize the Mean Squared Error (MSE) of the GNN model. The algorithm starts by initializing the current iteration \(t\) to 1, the patience count \(p\) to 1, and the best MSE (\(MSE_{\text{best}}\)) to infinity. The bounds for the hyperparameters \(k\) and \(l\) are also set at this stage.

The first step involves converting the 3D CAD dataset into a polygon mesh representation. This transformation is crucial as it prepares the CAD data for the subsequent re-meshing and graph representation processes.

The core of the optimization process is a while loop that continues until the maximum number of iterations (\(t_{\text{max}}\)) or the patience count (\(p_{\text{max}}\)) is reached. On the first iteration, the surrogate model of BO (such as a Gaussian Process) is initialized with the initial history \(\mathcal{H}\). In each iteration, the hyperparameters \((k_t, l_t)\) for that iteration are selected by maximizing the Expected Improvement (EI) acquisition function. This selection process identifies the hyperparameters that are expected to yield the most significant improvement in model performance.

Next, the mesh subdivision is performed using the selected \(k_t\) value. Clustering and re-meshing are then performed using the \(l_t\) value to adjust the mesh quality. The re-meshed mesh is then converted into a graph representation suitable for input into the GNN model. The weights of the GNN model are initialized, and the model is trained until early stopping criteria are met. This training process aims to minimize the MSE between the predicted and ground-truth values. After training, the MSE (\(MSE_t\)) of the GNN model for the current hyperparameters is computed.

If the computed \(MSE_t\) is less than the current best MSE (\(MSE_{\text{best}}\)), \(MSE_{\text{best}}\) is updated with \(MSE_t\), and the patience count \(p\) is reset to 1. If not, the patience count \(p\) is incremented by 1. The current hyperparameters and their corresponding MSE are then added to the history \(\mathcal{H}\), and the surrogate model is updated with this new history. The iteration counter \(t\) is then incremented by 1.

Finally, after completing the loop, the algorithm returns the hyperparameters \((k^*, l^*)\) that resulted in the minimum MSE recorded in the history \(\mathcal{H}\).

This BO approach, utilizing EI as the acquisition function, ensures a systematic and efficient search for the optimal mesh sizes, thereby enhancing the GNN model's prediction accuracy. A pseudo-code of the entire BO process used is illustrated in Algorithm. \ref{alg:BO-GNN}.

\begin{table*}[!h]
    \centering
    \caption{Comparison of 3D CNN, GCN, GNN, and BO-EI GNN models for prediction accuracy}
    \label{tab:Comparison-3DCNN-GCN-GNN}
    
    \resizebox{\linewidth}{!}{%
    \begin{tabularx}{\linewidth}{ccXXXXXXXXXX}
        
        % \toprule
        \multicolumn{11}{r}{*units: Mass (kg), Rim stiffness (kgf/mm), Disk stiffness (kgf/mm)} \\
        \toprule
        \multirow{2}{*}{Methods} & \multirow{2}{*}{\makecell[c]{Engineering \\ performance}} & \multicolumn{3}{c}{Training set} & \multicolumn{3}{c}{Validation set} & \multicolumn{3}{c}{Test set} \\ 
         & & RMSE & MAPE & $R^2$ & RMSE & MAPE & $R^2$ & RMSE & MAPE & $R^2$ \\ 
         \midrule
        
        \multirow{3}{*}{3D CNN} 
         & Mass & 0.09 & 0.41 & 0.991 & 0.17 & 0.69 & 0.952 & 0.28 & 1.22 & 0.915 \\ 
         & Rim stiffness & 240.31 & 1.74 & 0.992 & 269.42 & 1.73 & 0.973 & 488.44 & 2.77 & 0.931 \\ 
         & Disk stiffness & 1144.87 & 7.41 & 0.915 & 809.63 & 4.01 & 0.907 & 1409.8 & 6.85 & 0.728 \\ 
        \cmidrule{1-11}
        
        \multirow{3}{*}{GCN} 
         & Mass & 0.45 & 1.99 & 0.863 & 0.46 & 2.05 & 0.885 & 0.62 & 2.71 & 0.756 \\ 
         & Rim stiffness & 443.71 & 3.07 & 0.968 & 694.58 & 5.03 & 0.931 & 958.08 & 6.1 & 0.888 \\ 
         & Disk stiffness & 708.2 & 4.47 & 0.967 & 1182.86 & 8.19 & 0.924 & 1404.76 & 9.37 & 0.892 \\ 
        \cmidrule{1-11}
        
        \multirow{3}{*}{GNN} 
        & Mass & 0.19 & 0.87 & 0.982 & 0.38 & 1.62 & 0.924 & 0.41 & 1.8 & 0.894 \\ 
        & Rim stiffness & 226.17 & 1.57 & 0.993 & 690.66 & 4.7 & 0.931 & 802.06 & 5.06 & 0.923 \\ 
        & Disk stiffness & 785.11 & 4.91 & 0.962 & 1163.18 & 7.62 & 0.922 & 1488.39 & 9.91 & 0.876 \\ 
        \cmidrule{1-11}
        
        \multirow{3}{*}{\makecell[c]{\textbf{BO-EI GNN} \\ \textbf{(our proposed)}}} 
        & Mass & 0.07 & 0.29 & 0.997 & 0.16 & 0.67 & 0.986 & 0.16 & 0.69 & \textbf{0.985} \\ 
        & Rim stiffness & 200.82 & 1.36 & 0.994 & 208.93 & 1.42 & 0.993 & 350.13 & 2.25 & \textbf{0.978} \\ 
        & Disk stiffness & 494.79 & 3.13 & 0.985 & 593.79 & 3.49 & 0.979 & 758.03 & 4.94 & \textbf{0.963} \\ 
        
        \bottomrule
    \end{tabularx}
    }
\end{table*}
%%%%%%%%%%%% 3. Results and Discussion %%%%%%%%%%%%
\section{Results and Discussion}
\label{sec:results}
%
%%%%%%%%%%%% 3.1. Proposed Model Prediction Result %%%%%%%%%%%%
%
\subsection{Proposed Model Prediction Result} 
\label{subsec:model-predic}
When performing Bayesian Optimization (BO), three key considerations are necessary. The first is the choice of a prior function, optimized to approximate the GNN model serving as the surrogate model for BO. We selected the Gaussian process (GP) as a prior function because GP has high flexibility and tractability to express the assumption for data distribution. The second is an acquisition function that recommends evaluating the next set of hyperparameters. We selected the expected improvement (EI) instead of upper confidence bound (UCB), which is superior in function evaluations and time elapsed as the acquisition function for our proposed framework. Here, we refer to a GNN model with EI as “BO-EI GNN,” with UCB as “BO-UCB GNN,” and with the Metropolis–Hastings algorithm of Markov Chain Monte Carlo as “MCMC GNN.” The comparison of those models is discussed in Section \ref{subsec:comparison-different-models}. Finally, the bounds for $k$ and $l$, s.t. $\left\{k,l\in\mathbb{Z}\ |2\le k\le4,\ \ 3,000\le l\le5,000 \right\}$ were set in this study. 
\begin{equation}
    \left(R^2\right)^p=1-\ \frac{\sum_{i}^{n}\left(y_i^p-{\hat{y}}_i^p\right)^2}{\sum_{i}^{n}\left(y_i^p-{\bar{y}}_i^p\right)^2}\
    \label{eq:R2}
\end{equation}
\begin{equation}
    AE^p=\left|y_i^p-{\hat{y}}_i^p\right|,i=1,\ldots,n
    \label{eq:AE}
\end{equation}
\begin{equation}
    RMSE^p=\sqrt{\frac{\sum_{i}^{n}{(y_i^p-{\hat{y}}_i^p)}^2}{n}}
    \label{eq:RMSE}
\end{equation}
\begin{equation}
    MAPE^p=100\times\ \frac{1}{n}\sum_{i=1}^{n}\left|\frac{y_i^p-{\hat{y}}_i^p}{y_i^p}\right|
    \label{eq:MAPE}
\end{equation} 
where $n$ is the number of input data, $i$ is the $(i)$th input data, $y_i^p$ and ${\hat{y}}_i^p$ are the $(i)$th ground truth, and the $(i)$th predicted engineering performance, respectively.

We evaluate the model prediction accuracy for the train, validation and test set with the R-squared ($R^2$), the absolute error (AE), root mean square error (RMSE), and mean absolute percentage error (MAPE) calculated using Eq. \eqref{eq:R2}, \eqref{eq:AE}, \eqref{eq:RMSE}, \eqref{eq:MAPE}, respectively.  Table \ref{tab:Comparison-3DCNN-GCN-GNN} presents the RMSE, MAPE, and $R^2$ metrics for the proposed BO-EI GNN model across the training, validation, and test set. For mass, the metrics outperform those for other engineering performances due to the narrower variance in the ground-truth distribution. The mass data had the narrowest variance, from approximately 15 to 20. In contrast, the disk stiffness data had the widest variance, from approximately 5,000 to 22,500. Thus, the BO-EI GNN model for mass can be trained with higher accuracy than the disk stiffness model.

\begin{table}[!t]
    \centering
    \caption{Optimal size of mesh elements from our proposed BO-EI GNN model and $R^2$ for the test set}
    \label{tab:optimal-size}
    \begin{tabularx}{0.6\linewidth}{cccc}
    \toprule
        \makecell[c]{Optimal size of mesh elements} & Mass & \makecell[c]{Rim \\ stiffness} & \makecell[c]{Disk \\ stiffness} \\ 
        \midrule
        \makecell[c]{Number of subdivision ($k$)} & 3 & 3 & 3 \\ 
        \makecell[c]{Number of clustering ($l$)} & 4,557 & 4,626 & 3,438 \\
        $R^2$ for the test set& 0.985 & 0.978 & 0.963 \\
    \bottomrule
    \end{tabularx}
\end{table}
\begin{figure}[h]
    \centering
    \includegraphics[width=0.7\linewidth]{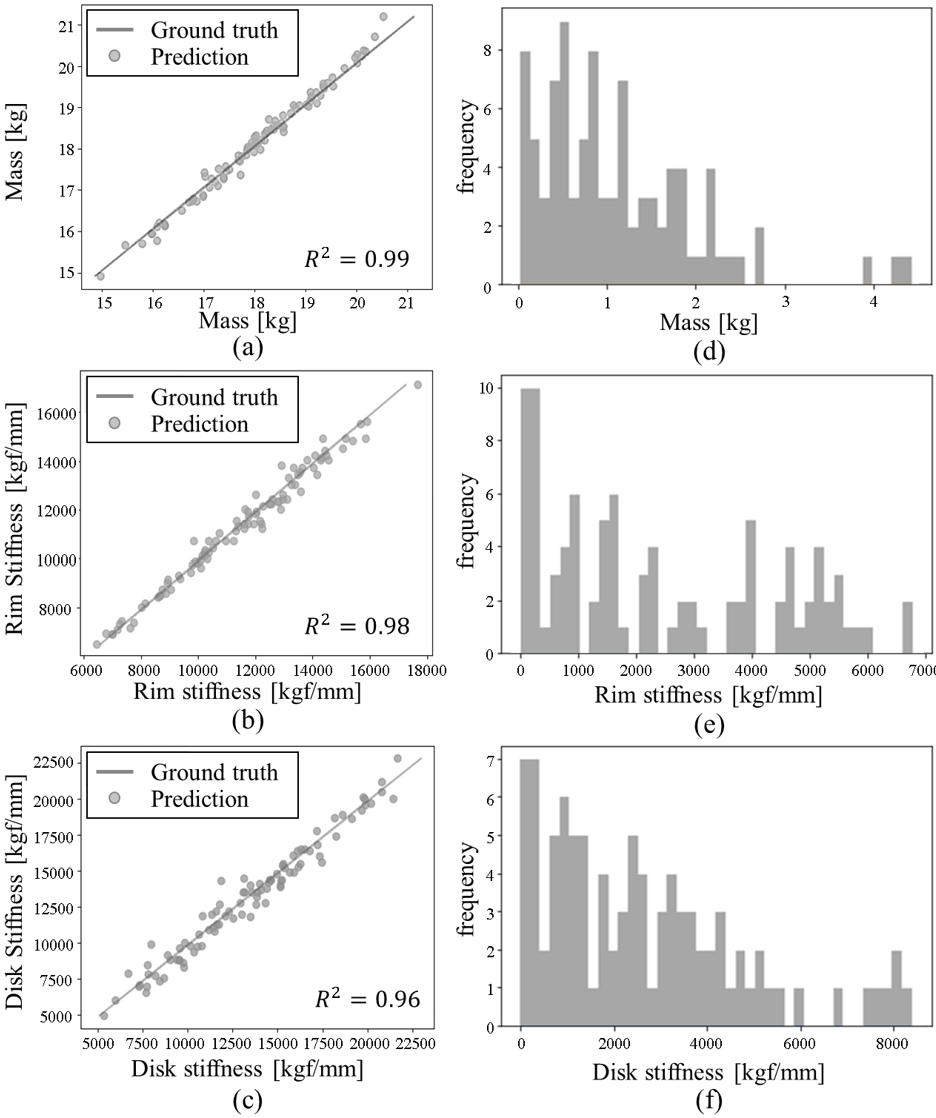}
    \caption{(a)-(c) Scatter plots show the relationships between BO-EI GNN model prediction and the ground truth for the test set and (d)-(f) Histogram of absolute error (AE) for the test set}
    \label{fig:scattor}
\end{figure}
Table \ref{tab:optimal-size} displays the optimal mesh element sizes for each engineering performance.
% All engineering performances with the BO-EI GNN model were evaluated with R-squared ($R^2$) calculated using Eq. \eqref{eq:R2}.
In the case of mass as an engineering performance, the $k$ and $l$ were derived with values of 3 and 4,557, respectively. Finally, the $R^2$ value of 0.985 was higher than that of the other engineering performance.
%

%
% $R^2$ indicates the proportion of variance in the independent variable (i.e., the engineering performance in this study) that is predictable from the independent variables.
Figure \ref{fig:scattor} (a)-(c) shows scatter plots comparing the predicted and actual values for each engineering performance on the test set. The solid line indicates the ground truth, and the dots indicate the predicted values. The dots of the mass closest to the solid line had the highest $R^2$ value. The AE is the error between the predicted and actual values for each engineering performance. The error distribution for the test set is shown in Fig. \ref{fig:scattor} (d)-(f).

%%%%%%%%%%%% 3.2. Comparison in Different Models %%%%%%%%%%%%
%
\subsection{Comparison in Different Models}
\label{subsec:comparison-different-models}
%
%%%%%%%%%%%% 3.2.1. Comparison of Re-meshing and Not Re-meshing %%%%%%%%%%%%

\subsubsection*{Comparison of Re-meshing and Not Re-meshing}
This subsection compares our proposed BO-EI GNN model with other models that do not utilize re-meshing algorithms, including those using the original polygon mesh dataset and voxel representation. We refer to the GNN model with Expected Improvement as “BO-EI GNN,” with Upper Confidence Bound as “BO-UCB GNN,” and with the Metropolis-Hastings algorithm of Markov Chain Monte Carlo as “MCMC GNN.”

In this section, the 3D CAD dataset was converted into different 3D data representations. Then, 3D deep learning models suitable for each representation were trained, evaluated, and compared without re-meshing algorithms, except for the BO-EI GNN model. Each representation proceeds with a voxel, point cloud, and graph representation. In this study, the point cloud is used with the PointNet model \cite{qi2017pointnet} by converting the classification prediction model into a regression model to predict the engineering performance. The results of the point cloud model were excluded due to its low prediction accuracy.

The voxel model was adopted for the 3D CNN model proposed by Maturana and Scherer \cite{maturana2015voxnet}, which predicts the engineering performance mass for 3D printing. After converting 3D CAD data into polygon mesh data, elements are expressed as occupied (value of 1) or non-occupied (value of 0) values by checking the connection information of the adjacent mesh nodes. The 3D CNN model used a voxel size of 64 × 64 × 64, three convolutional layers, and three fully connected layers. The three convolutional layers had 64, 16, and 8 filters, and the same padding was applied. The dimensions of the dense layers were the same (500, 200, and 25) as those in the BO-EI GNN model in this study. Except for the last layer, a rectified linear unit (ReLU) was used as the activation function for the remaining layers. Through a sufficient hyperparameter search, the 3D CNN model used the Adam optimizer, and the learning rate, epochs, patience, and batch size were set as 0.0001, 10,000, 100, and 8, respectively. The 3D CNN model evaluation for each engineering performance is presented in Table~\ref{tab:Comparison-3DCNN-GCN-GNN}.

We evaluated three models using graph representation: GNN, GCN, and BO-EI GNN. As explained in Section \ref{subsec:stage2}, the GNN model uses Eq. \eqref{eq:GNN} to update node features through message passing of the graph layers. The GCN model uses Eq. \eqref{eq:GCN} to perform graph convolutional operations. Except for having a different form of the adjacency matrices between the GNN and GCN models, both models used the same values and functions, such as the number of layers, the size of the latent vector, hyperparameters, activation functions, and an optimizer. Both models are trained with a polygon mesh dataset, which is not the optimal size of mesh elements. The results of the experiment show that BO-EI GNN model is excellent in terms of all metrics for each engineering performance, as shown in Table~\ref{tab:Comparison-3DCNN-GCN-GNN}. 

\begin{figure}[!h]
    \centering\includegraphics[width=0.8\linewidth]{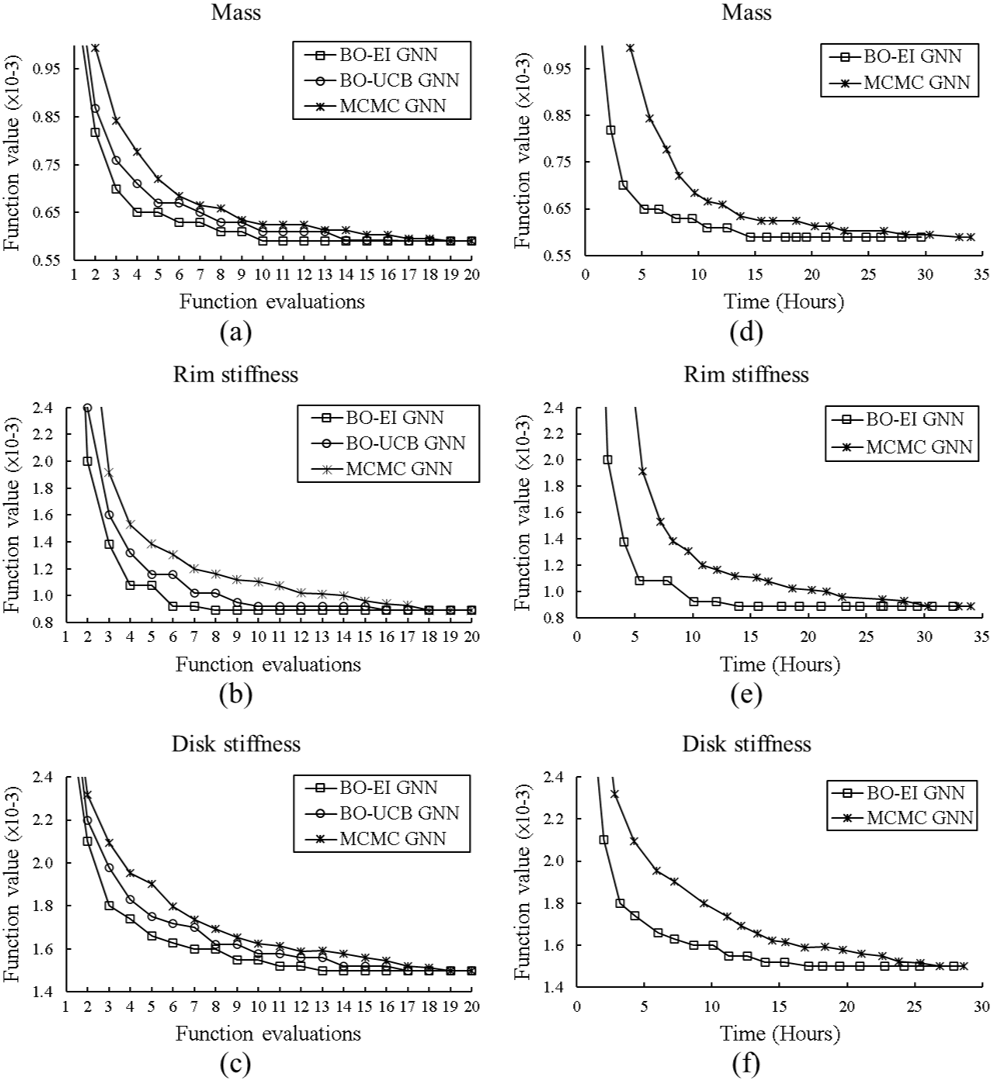}
    \caption{(a)-(c) Different strategies of optimization in terms of function evaluations and (d)-(f) Comparisons of BO-EI GNN and MCMC GNN models in terms of time elapsed with training for the validation test}
    \label{fig:different-strategies-func-eval}
\end{figure}
\begin{figure*}[!h]
    \centering\includegraphics[width=0.9\linewidth]{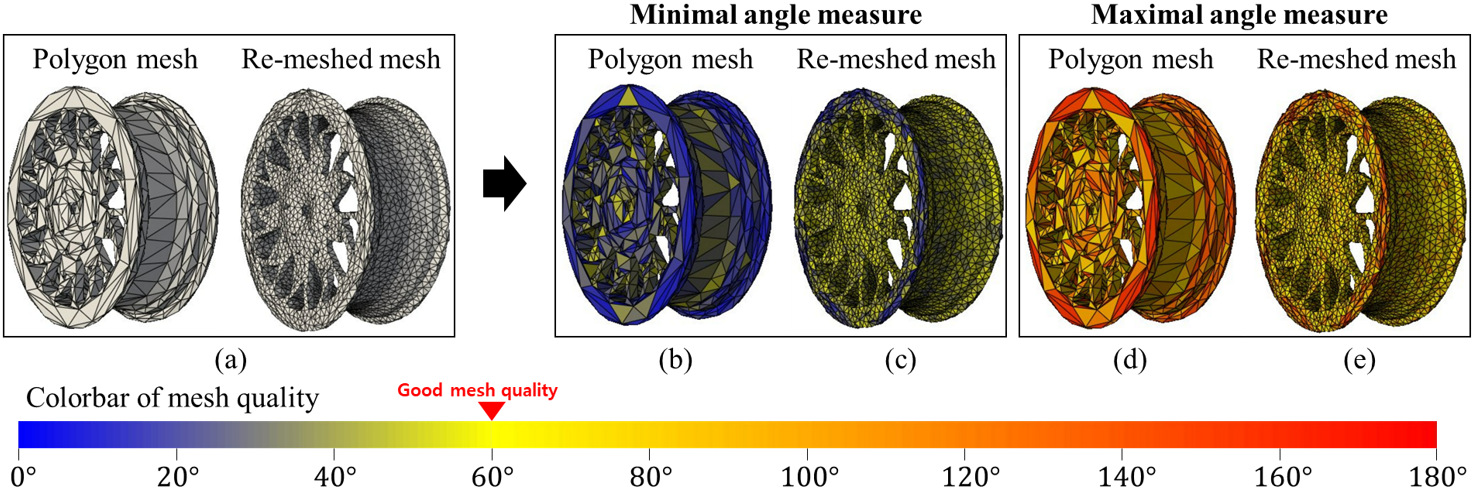}
    \caption{Visualization of mesh quality for minimal and maximal angle measures. (a) a polygon mesh converted from a 3D CAD, (b)-(c) in case of minimal angle measure and (d)-(e) in case of maximal angle measure}
    \label{fig:vis-mesh-qual}
\end{figure*}
\begin{figure}[!h]
    \centering\includegraphics[width=0.7\linewidth]{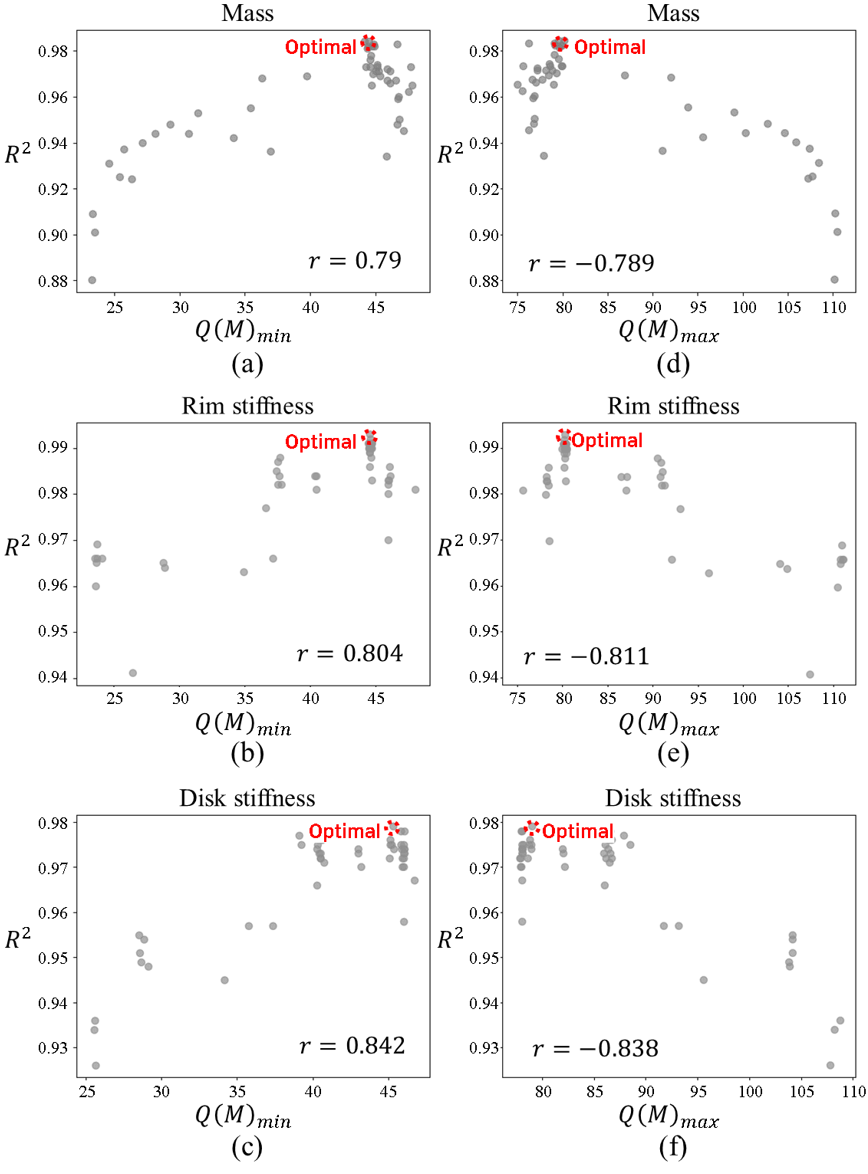}
    \caption{Pearson correlation coefficients and scatter plots of relationships between $R^2$ and mesh quality $Q(\mathcal{M})$. (a)-(c) In case of minimal angle measure and (d)-(f) In case of maximal angle measure.}
    \label{fig:pearson}
\end{figure}

%%%%%%%%%%%% 3.2.2. Comparison of BO-EI GNN, BO-UCB GNN, and MCMC GNN Models %%%%%%%%%%%%

\subsubsection*{Comparison of BO-EI GNN, BO-UCB GNN, and MCMC GNN Models} 

We empirically analyzed and compared the optimization strategies in terms of function evaluations and time elapsed in the BO-EI GNN, BO-UCB GNN, and MCMC GNN. We utilize the original graph dataset converted from the polygon mesh dataset as input dataset. In Fig. \ref{fig:different-strategies-func-eval}, “function value” indicates the values of MSE evaluated by each trained model. For “function evaluations,” MCMC GNN model training was performed 10 times for each number of function evaluations, and the mean value was reported. The MCMC GNN model can be evaluated in parallel, in contrast, the other models were run once because they should be trained by sequential iteration steps owing to the Bayesian optimization. The results of the analyses are presented in Fig. \ref{fig:different-strategies-func-eval} (a)-(c) in terms of function evaluations. For all the engineering performance, the BO-EI GNN model is superior to the BO-UCB GNN and significantly outperforms the MCMC GNN, with a minimum MSE of less than half as function evaluations and time costs using one GPU (Geforce RTX 3090) and CPU 64 Cores (four AMD EPYC 7282 16-core processors). In terms of time elapsed, the BO-EI model is compared with the MCMC GNN in Fig. \ref{fig:different-strategies-func-eval} (d)-(f). 

%%%%%%%%%%%% 3.3. Comparison in Different Models %%%%%%%%%%%%
%

 High-quality meshes are crucial for practical applications such as 3D visualization, numerical simulations, animation, and 3D deep learning-based surrogate models. We discuss mesh quality measures with the visualization of the optimal size of the mesh elements and  examine the relationship between the mesh quality values using two measures and the values of the model prediction accuracy.

%%%%%%%%%%%% 3.3.1 Mesh Quality Measures %%%%%%%%%%%%

\subsubsection*{Mesh Quality Measures}

In finite element analysis (FEA) and computer graphics, there are various measures for evaluating the quality of a mesh, including the minimal angle, maximal angle, aspect ratio, regularity, and feature preservation. This study assesses mesh quality using minimal and maximal angle measures. We evaluated the mesh quality for one mesh dataset $\mathcal{M}\equiv\left\{M_1,M_2\ldots M_{925}\right\}$ by applying the minimal and maximal angle measures of $Q\left(\mathcal{M}\right)_{min}$ and $Q\left(\mathcal{M}\right)_{max}$, respectively. The equations for mesh quality are as follows:
\begin{equation}
    Q\left(\mathcal{M}_{k,l}\right)_{min}=\frac{1}{n}\sum_{i=1}^{n}{\frac{1}{s}\sum_{j=1}^{s}min\left[\left\{\theta\ \right|\ \theta\in{M_i(\phi}_j)\}\right]}
    \label{eq:Q_min}
\end{equation}
\begin{equation}
    Q\left(\mathcal{M}_{k,l}\right)_{max}=\frac{1}{n}\sum_{i=1}^{n}{\frac{1}{s}\sum_{j=1}^{s}max\left[\left\{\theta\ \right|\ \theta\in{M_i(\phi}_j)\}\right]}
    \label{eq:Q_max}
\end{equation}
where $\theta$ is the interior angle of the $(j)$th triangle cell $\phi_j\in M_i$, $s$ is the number of triangular cells $\phi_j$, and $n$ is the number of mesh data $M_i\in\mathcal{M}$. One mesh data $M_i$ of mesh dataset $\mathcal{M}_{k,l}$ consists of $s$ cells $\phi_j$. One cell $\phi_j$ includes the values of the three angles $\theta\in M_i(\phi_j)$ because this study uses a cell as a triangle, which means that the sum of the three angles is 180°. Ultimately, the minimal angle measure $Q\left(\mathcal{M}_{k,l}\right)_{min}$ is calculated using Eq. \eqref{eq:Q_min} and which is the summation of the smallest angles among the three angles. 

For example, for mass, the re-meshed mesh dataset with optimal mesh element sizes is denoted as $\mathcal{M}{3,4557}$, where $k$ and $l$ are 3 and 4,557, respectively, as shown in Table \ref{tab:optimal-size} 
In other words, $\mathcal{M}_{3,4557}$, which is clustered into 4557 and divided three times. Figure \ref{fig:vis-mesh-qual} shows the visualization with minimal and maximal angle measures for an arbitrary 585th mesh dataset $M_{585}$ of mesh dataset $\mathcal{M}_{3,4557}$. Generally, mesh quality improves as the triangle cells approach the shape of an equilateral triangle. 

In the case of a polygon mesh converted from a 3D CAD in Fig. \ref{fig:vis-mesh-qual} (a), the colors of the cells are expressed more as blue or red, as shown in Fig. \ref{fig:vis-mesh-qual} (b) and (d). In contrast, the re-meshed meshes in yellow are distributed more than those in blue (or red part), which indicates closer to 60 °, as shown in Fig. \ref{fig:vis-mesh-qual} (c) and (d). Thus, we can verify that the optimal size of the mesh elements is higher than that of a polygon mesh.
%
%

%%%%%%%%%%%% 3.3.2. Relationship between Mesh Quality and Model Prediction Accuracy %%%%%%%%%%%%

\subsection{Mesh Quality}
\label{subsec:mesh-quality}

\subsubsection*{Relationship between Mesh Quality and Model Prediction Accuracy}
We generated 50 re-meshed mesh datasets $\left\{\mathcal{M}_1,\ \mathcal{M}_2\ldots\mathcal{M}_{50}\right\}$ by randomly sampling $k$ and $l$ values, which determine the mesh quality. One of these 50 samples contains the optimal mesh element sizes obtained in Section \ref{subsec:model-predic}. In addition, we derived 50 $R^2$ scores and the values of the model prediction accuracy by training the GNN model on 50 dataset. 

We used the Pearson correlation coefficient, $r$, to measure linear correlation. $r$ is a number between -1 and 1, which measures the strength and direction of the relationship between the two variables. The Pearson correlation coefficient calculated using Eq. \eqref{eq:coef} was employed for a total of six cases with a combination of three types of the engineering performance and two types of mesh quality measures (minimal and maximal angle), as shown in Fig. \ref{fig:pearson}.

\begin{equation}
    r^p= \frac{\sum_{i=1}^{m}[Q(\mathcal{M})_i-\bar{Q}(\mathcal{M})_i][(R^2)_i^p-(\bar{R^2})_i^p]}
    {\sqrt{\sum_{i=1}^{m}[Q(\mathcal{M})_i-{\bar{Q}(\mathcal{M})_i}]^2\sum_{i=1}^{m}[(R^2)_i^p-(\bar{R^2})_i^p]^2}}\
    \label{eq:coef}
\end{equation}
where $m$ is the number of data samples (where m is 50) and $i$ is the number of orders for both the mesh quality of the $\left(i\right)$th re-meshed dataset and the prediction accuracy of the $(i)$th GNN model out of 50 samples.
In the case of the minimal angle measure shown in Fig. \ref{fig:pearson} (a)-(c), the Pearson correlation coefficient $r$ has values of 0.79, 0.804, and 0.842 for mass, rim stiffness, and disk stiffness, respectively. The minimal angle measure and model prediction accuracy have a high positive correlation because $r$ is above 0.7, which indicates a high correlation. Model prediction accuracy generally improved with higher mesh quality. However, the optimal point of the mesh elements with the highest prediction accuracy (red dashed circles in Fig. \ref{fig:pearson} (a)-(c) was not located in the upper right corner. Figure \ref{fig:pearson} (d)-(f) shows the results of the maximal angle measure, and each r value shows high negative correlations of -0.789, -0.811, and -0.838 for each engineering performance. Equivalent to the minimal measures, the optimal point that improves the model accuracy obtained through Bayesian optimization is not located in the upper-left corner. Although the model accuracy is highly related to the mesh quality, it is necessary to derive the optimal mesh element size.
%
%%%%%%%%%%%% 4. Conclusion %%%%%%%%%%%%
%

\section{Conclusion}
\label{sec:conclusion}

In this study, we introduced a Bayesian Graph Neural Network (GNN) framework designed to optimize 3D deep learning-based surrogate models for predicting engineering performance from 3D CAD datasets. Our framework effectively addresses the inherent challenges of dealing with complex 3D geometries and computationally intensive simulations. By leveraging Bayesian optimization, we determined the optimal size of mesh elements, striking a balance between prediction accuracy and computational efficiency.

Our experimental results demonstrated the following key points:
\begin{enumerate}
    \item \textbf{High-Accuracy Predictions:} The proposed BO-EI GNN model achieved superior prediction accuracy for engineering performance metrics such as mass, rim stiffness, and disk stiffness compared to other models, including 3D CNN, GCN, and standard GNN models. This was evidenced by consistently lower RMSE and MAPE values, and higher $R^2$ scores across training, validation, and test sets.

     \item \textbf{Optimal Mesh Element Sizes:} Through Bayesian optimization, we identified optimal mesh element sizes, which significantly improved the prediction accuracy of the surrogate models. The optimal sizes for subdivisions ($k$) and clustering ($l$) were specific to each engineering performance, as highlighted by our results.

     \item \textbf{Impact of Mesh Quality on Model Accuracy:} Our analysis showed a strong correlation between mesh quality and model prediction accuracy. High-quality meshes, characterized by angles closer to those of an equilateral triangle, contributed to better predictive performance. This finding underscores the importance of mesh preprocessing in developing accurate surrogate models.

     \item \textbf{Efficiency of Bayesian Optimization:} Compared to traditional methods like the Metropolis-Hastings algorithm of Markov Chain Monte Carlo (MCMC), our Bayesian optimization approach (using Expected Improvement as the acquisition function) demonstrated greater efficiency in terms of function evaluations and computational time. This efficiency makes our framework particularly suitable for iterative design processes and complex engineering analyses.

     \item \textbf{Applicability Across Engineering Disciplines:} Our proposed framework aligns graph representation with mesh representation, which is structurally identical to the mesh representation used in the Finite Element Method (FEM). This structural similarity facilitates the future use of physical information derived from solving complex simulations. This capability enhances the framework’s ability to incorporate physics-based information, adapt to different simulation types, and thereby significantly increase the utility of surrogate models in computer-aided engineering (CAE).
\end{enumerate}

By addressing key challenges such as memory efficiency, ease of rendering, and handling varying input sizes, our framework represents a significant advancement in the development of surrogate models for complex engineering simulations. The integration of Bayesian optimization ensures that these models are both accurate and computationally feasible, providing a robust solution to the limitations of existing methods.

Future work could focus on extending this framework to other types of engineering simulations and further exploring the integration of physical information derived from solving partial differential equations (PDEs). Additionally, investigating the application of our framework in real-time scenarios and large-scale industrial datasets could provide valuable insights into its practical utility and scalability.

In conclusion, our proposed Bayesian GNN framework offers a powerful and efficient approach for enhancing engineering performance prediction, potentially transforming the way complex simulations are conducted in various engineering disciplines.

\section*{Acknowledgments}
This work was supported by the National Research Foundation of Korea grant (2018R1A5A7025409) and the Ministry of Science and ICT of Korea grant (No.2022-0-00969, No.2022-0-00986).

\section*{Conflict of interest statement}
The authors declared no potential conflicts of interest.

\section*{Replication of results}
The codes will be available upon request.

%Bibliography
\bibliographystyle{unsrt}  
\bibliography{references}

\begin{thebibliography}{10}

\bibitem{cunningham2019investigation}
James~D Cunningham, Timothy~W Simpson, and Conrad~S Tucker.
\newblock An investigation of surrogate models for efficient performance-based decoding of 3d point clouds.
\newblock {\em Journal of Mechanical Design}, 141(12), 2019.

\bibitem{alizadeh2020managing}
Afshin Alizadeh and Others.
\newblock Managing the computational cost of surrogate models in engineering design optimization.
\newblock {\em Computers \& Industrial Engineering}, 139:105703, 2020.

\bibitem{wu20153d}
Zhirong Wu, Shuran Song, Aditya Khosla, Fisher Yu, Linguang Zhang, Xiaoou Tang, and Jianxiong Xiao.
\newblock 3d shapenets: A deep representation for volumetric shapes.
\newblock In {\em Proceedings of the IEEE conference on computer vision and pattern recognition}, pages 1912--1920, 2015.

\bibitem{hanocka2019meshcnn}
Rana Hanocka, Amir Hertz, Noa Fish, Raja Giryes, Shachar Fleishman, and Daniel Cohen-Or.
\newblock Meshcnn: a network with an edge.
\newblock {\em ACM Transactions on Graphics (TOG)}, 38(4):1--12, 2019.

\bibitem{agathos20103d}
Alexander Agathos, Ioannis Pratikakis, Panagiotis Papadakis, Stavros Perantonis, Philip Azariadis, and Nickolas~S Sapidis.
\newblock 3d articulated object retrieval using a graph-based representation.
\newblock {\em The Visual Computer}, 26:1301--1319, 2010.

\bibitem{qi2017pointnet}
Charles~R Qi, Hao Su, Kaichun Mo, and Leonidas~J Guibas.
\newblock Pointnet: Deep learning on point sets for 3d classification and segmentation.
\newblock In {\em Proceedings of the IEEE conference on computer vision and pattern recognition}, pages 652--660, 2017.

\bibitem{qi2017pointnet++}
Charles~Ruizhongtai Qi, Li~Yi, Hao Su, and Leonidas~J Guibas.
\newblock Pointnet++: Deep hierarchical feature learning on point sets in a metric space.
\newblock {\em Advances in neural information processing systems}, 30, 2017.

\bibitem{xiang2015data}
Yu~Xiang, Wongun Choi, Yuanqing Lin, and Silvio Savarese.
\newblock Data-driven 3d voxel patterns for object category recognition.
\newblock In {\em Proceedings of the IEEE conference on computer vision and pattern recognition}, pages 1903--1911, 2015.

\bibitem{maturana2015voxnet}
Daniel Maturana and Sebastian Scherer.
\newblock Voxnet: A 3d convolutional neural network for real-time object recognition.
\newblock In {\em 2015 IEEE/RSJ international conference on intelligent robots and systems (IROS)}, pages 922--928. IEEE, 2015.

\bibitem{feng2019meshnet}
Yutong Feng, Yifan Feng, Haoxuan You, Xibin Zhao, and Yue Gao.
\newblock Meshnet: Mesh neural network for 3d shape representation.
\newblock In {\em Proceedings of the AAAI conference on artificial intelligence}, volume~33, pages 8279--8286, 2019.

\bibitem{tatarchenko2017octree}
Maxim Tatarchenko, Alexey Dosovitskiy, and Thomas Brox.
\newblock Octree generating networks: Efficient convolutional architectures for high-resolution 3d outputs.
\newblock In {\em Proceedings of the IEEE international conference on computer vision}, pages 2088--2096, 2017.

\bibitem{wang2017cnn}
Peng-Shuai Wang, Yang Liu, Yu-Xiao Guo, Chun-Yu Sun, and Xin Tong.
\newblock O-cnn: Octree-based convolutional neural networks for 3d shape analysis.
\newblock {\em ACM Transactions On Graphics (TOG)}, 36(4):1--11, 2017.

\bibitem{williams2019design}
Glen Williams, Nicholas~A Meisel, Timothy~W Simpson, and Christopher McComb.
\newblock Design repository effectiveness for 3d convolutional neural networks: Application to additive manufacturing.
\newblock {\em Journal of Mechanical Design}, 141(11), 2019.

\bibitem{yoo2021explainable}
Soyoung Yoo and Namwoo Kang.
\newblock Explainable artificial intelligence for manufacturing cost estimation and machining feature visualization.
\newblock {\em Expert Systems with Applications}, 183:115430, 2021.

\bibitem{shin2023wheel}
Seungyeon Shin, Ah-hyeon Jin, Soyoung Yoo, Sunghee Lee, ChangGon Kim, Sungpil Heo, and Namwoo Kang.
\newblock Wheel impact test by deep learning: prediction of location and magnitude of maximum stress.
\newblock {\em Structural and Multidisciplinary Optimization}, 66(1):24, 2023.

\bibitem{scarselli2008graph}
Franco Scarselli, Marco Gori, Ah~Chung Tsoi, Markus Hagenbuchner, and Gabriele Monfardini.
\newblock The graph neural network model.
\newblock {\em IEEE transactions on neural networks}, 20(1):61--80, 2008.

\bibitem{markovitch2021enhancing}
Evgeniy Markovitch and Marjolein Fokkema.
\newblock Enhancing prediction rule ensembles through model-based data generation.
\newblock {\em Machine Learning}, 110:1247--1267, 2021.

\bibitem{balaprakash2013iterative}
Prasanna Balaprakash, Stefan~M Wild, and Paul~D Hovland.
\newblock An iterative parallel algorithm for high-performance data analytics.
\newblock In {\em 2013 IEEE International Conference on Big Data}, pages 74--81. IEEE, 2013.

\bibitem{granacher2021improvement}
Franziska Granacher and Others.
\newblock Improvement of surrogate model accuracy through active learning with continuous data enrichment.
\newblock {\em Engineering Applications of Artificial Intelligence}, 97:104019, 2021.

\bibitem{jones2023data}
Nathan Jones and Others.
\newblock Data augmentation and fine-tuning for pre-trained models in engineering applications.
\newblock {\em Advanced Engineering Informatics}, 49:101395, 2023.

\bibitem{cao2020graph}
Yanwei Cao, Jian Bai, Yuan Tang, and Others.
\newblock Graph representation learning for machining feature recognition.
\newblock {\em Journal of Manufacturing Systems}, 57:293--304, 2020.

\bibitem{yoo2021integrating}
Soyoung Yoo, Sunghee Lee, Seongsin Kim, Kwang~Hyeon Hwang, Jong~Ho Park, and Namwoo Kang.
\newblock Integrating deep learning into cad/cae system: generative design and evaluation of 3d conceptual wheel.
\newblock {\em Structural and multidisciplinary optimization}, 64(4):2725--2747, 2021.

\bibitem{AtairSimLab}
Inc. Atair~Aerospace.
\newblock Simlab.
\newblock \url{https://www.atair.com/simlab/}, accessed March 12, 2023.

\bibitem{Zhou2018}
Qian-Yi Zhou, Jaesik Park, and Vladlen Koltun.
\newblock Open3d: A modern library for 3d data processing.
\newblock {\em arXiv:1801.09847}, 2018.

\bibitem{valette2008generic}
S{\'e}bastien Valette, Jean~Marc Chassery, and R{\'e}my Prost.
\newblock Generic remeshing of 3d triangular meshes with metric-dependent discrete voronoi diagrams.
\newblock {\em IEEE Transactions on Visualization and Computer Graphics}, 14(2):369--381, 2008.

\bibitem{khan2020surface}
Dawar Khan, Alexander Plopski, Yuichiro Fujimoto, Masayuki Kanbara, Gul Jabeen, Yongjie~Jessica Zhang, Xiaopeng Zhang, and Hirokazu Kato.
\newblock Surface remeshing: A systematic literature review of methods and research directions.
\newblock {\em IEEE transactions on visualization and computer graphics}, 28(3):1680--1713, 2020.

\bibitem{aurenhammer2013voronoi}
Franz Aurenhammer, Rolf Klein, and Der-Tsai Lee.
\newblock {\em Voronoi diagrams and Delaunay triangulations}.
\newblock World Scientific Publishing Company, 2013.

\bibitem{fey2018splinecnn}
Matthias Fey, Jan~Eric Lenssen, Frank Weichert, and Heinrich M{\"u}ller.
\newblock Splinecnn: Fast geometric deep learning with continuous b-spline kernels.
\newblock In {\em Proceedings of the IEEE conference on computer vision and pattern recognition}, pages 869--877, 2018.

\bibitem{bianchi2020spectral}
Filippo~Maria Bianchi, Daniele Grattarola, and Cesare Alippi.
\newblock Spectral clustering with graph neural networks for graph pooling.
\newblock In {\em International conference on machine learning}, pages 874--883. PMLR, 2020.

\bibitem{snoek2012practical}
Jasper Snoek, Hugo Larochelle, and Ryan~P Adams.
\newblock Practical bayesian optimization of machine learning algorithms.
\newblock {\em Advances in neural information processing systems}, 25, 2012.

\end{thebibliography}

\end{document}